\documentclass{article}

    \PassOptionsToPackage{numbers, compress}{natbib}

\usepackage[preprint]{neurips_2022_arxiv}

\usepackage[utf8]{inputenc} 
\usepackage[T1]{fontenc}    
\usepackage{hyperref}       
\usepackage{url}            
\usepackage{booktabs}       
\usepackage{amsfonts}       
\usepackage{nicefrac}       
\usepackage{microtype}      
\usepackage{color,xcolor}         

\usepackage{subfig,graphicx}
\usepackage{multirow}
\usepackage{adjustbox}
\usepackage{amstext}
\usepackage{amssymb,amsmath}
\usepackage{xspace}
\usepackage{bm}
\usepackage{bbm}
\usepackage{wrapfig}
\usepackage{pifont}

\def \alambic {\includegraphics[width=6.5pt]{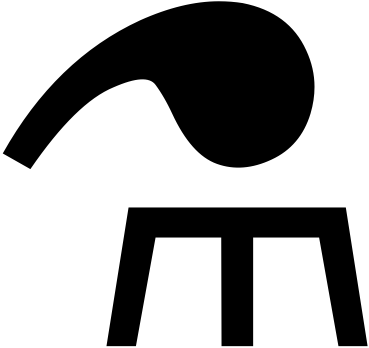}\xspace}
\def \cmk {\ding{51}}
\def \xmk {\ding{55}}

\title{SemFormer: Semantic Guided Activation Transformer for Weakly Supervised Semantic Segmentation}

\author{%
    Junliang Chen, Xiaodong Zhao, Cheng Luo, Linlin Shen\thanks{Corresponding Author}\\
  Shenzhen University \\
  \texttt{\{chenjunliang2016,zhaoxiaodong2020,luocheng2020\}@email.szu.edu.cn} \\
  \texttt{llshen@szu.edu.cn}
}

\begin{document}

\maketitle

\begin{abstract}

Recent mainstream weakly supervised semantic segmentation (WSSS) approaches are mainly based on Class Activation Map (CAM) generated by a CNN (Convolutional Neural Network) based image classifier. In this paper, we propose a novel transformer-based framework, named Semantic Guided Activation Transformer (SemFormer), for WSSS. We design a transformer-based Class-Aware AutoEncoder (CAAE) to extract the class embeddings for the input image and learn class semantics for all classes of the dataset. The class embeddings and learned class semantics are then used to guide the generation of activation maps with four losses, \emph{i.e.}, class-foreground, class-background, activation suppression, and activation complementation loss. Experimental results show that our SemFormer achieves \textbf{74.3}\% mIoU and surpasses many recent mainstream WSSS approaches by a large margin on PASCAL VOC 2012 dataset. Code will be available at \url{https://github.com/JLChen-C/SemFormer}.
\end{abstract}

\section{Introduction}

Semantic Segmentation, a fundamental but challenging task in computer vision, has received much attention of researchers. This task is to assign a class for each pixel in the given image. Due to the rapid development of computer vision community, many approaches \cite{fcn,deeplabv1,deeplabv2,bisenet,setr,segformer} have achieved surprising performance for fully-supervised semantic segmentation (FSSS). However, FSSS is developed upon pixel-level annotations, which is very time-consuming and may cost huge human labour. To reduce the numerous resource cost, recently many weakly supervised semantic segmentation (WSSS) approaches aims at making full use of weaker annotations, \emph{e.g.}, bounding boxes \cite{bcm, sdi}, scribbles \cite{rawks,scribblesup}, points \cite{whatpoint}, and image-level labels \cite{oaa,aepsl,cpn,gain,ecsnet}. These WSSS approaches follow a two-stage paradigm: first generating pseudo semantic segmentation labels utilizing the weak annotations, and then training an FSSS network via the pseudo labels. In the above approaches, image-level annotations are the most convenient labels to acquire, and are therefore broadly studied by researchers. As a result, in this paper, we mainly pay attention to approaches using image-level annotations.

However, the image-level labels can not provide enough location information. Fortunately, the development of Class Activation Map (CAM) \cite{cam} provides an effective way to acquire the location information only using image-level annotations. Many WSSS approaches are built upon CAM. Although CAM is simple and effective, it has an obvious disadvantage, \emph{i.e.}, it only activates the most discriminative regions, which results in under-activation. As a result, many WSSS approaches including CAM expanding \cite{oaa,mdc,seam} are trying to solve this problem. Erasing-based methods drive the network to focus on the less discriminative regions by erasing the regions with high response \cite{gain,aepsl,occse}.

Recently, vision transformers has been introduced to computer vision community. Vision Transformer (ViT) \cite{vit} splits the given image into patch sequence. The patches are projected to patch embeddings by a shared fully-connected (FC) layer and summed up with corresponding learnable position embeddings. The summation is then input to the sequential transformer blocks which contain Multi-Head Self-Attention (MHSA) and Multi-Layer Perceptron (MLP). Thanks to the MHSA and MLP, ViT can build long-range relationship between the patches and handle complex feature transforms. Recent researches \cite{vit,deit,pvt,swin} show the superiority of vision transformers over conventional convolutional neural network (CNN).

In this paper, we propose a novel framework with vision transformer for WSSS, which generates activation maps based on class semantics. In this framework, we design a transformer based Class-Aware AutoEncoder (CAAE) to extract class embeddings of the input image and learn semantic of each class in the dataset with the semantic similarity loss. Via the CAAE and the learned class semantics, we design an adversarial learning framework for generating class activation map (CAM) for each class using transformer. Given an input image and its CAM, we construct an complementary image pair, class-foreground and class-background images, by multiplying the CAM with the given image. Both of the complementary images are input to the CAAE to extract the corresponding class embedding. While the similarity between the embedding of the class-foreground image and the class semantic should be maximized, the similarity between the embedding of the class-background image and the corresponding class semantic should be minimized. We design two losses, activation suppression, and activation complementary loss, to improve the activation map. Besides, we introduce extra class tokens into the segmentation network to learn the class embeddings, and refine the CAM using the class-relative attention of each class.

We summarize our contribution as follows. First, different from conventional classification network based CAM approaches, we design a Class-Aware AutoEncoder (CAAE) to extract the class embedding of the foreground and background images and learn the semantic of each class in the dataset, to guide the generation of the activation map. Second, to maximize the similarity between the class embedding of the input image and the semantic of this class, adversarial losses are designed to measure the similarity between the class-foreground or class-background image and this class, and thus guide the generation of the activation map. Third, we introduce extra class tokens into the segmentation network to learn the class embeddings and refine the CAM with the generated multi-head self attention of each class. Finally, experimental results show that our proposed SemFormer achieves \textbf{74.3}\% mIoU and surpasses many recent mainstream WSSS approaches on PASCAL VOC 2012 dataset.

\vspace{-12pt}
\section{Related Work}
\vspace{-12pt}

\subsection{Weakly-Supervised Semantic Segmentation}
\vspace{-5pt}

As CAM only activates the most discriminative regions of the given image, many approaches attempt to address this issue. \cite{dsrg,sec} use object boundary information to expand the regions of the initial CAM. \cite{cian, affinitynet} utilize the pixel relationship to refine the initial CAM to generate better pseudo labels. But these approaches heavily rely on the quality of the initial CAM. Many methods try to improve the quality of the initial CAM. MDC \cite{mdc} introduce multi-dilated convolutions to enlarge the receptive field for classification of the non-discriminative regions. SEAM \cite{seam} designs a module exploring the pixel context correlation. However, these methods bring complicated modules to the segmentation network or require complex training procedure. Beyond above approaches, another popular technique, adversarial erasing (AE), generally erases the discriminative regions found by CAM, and drives the network to recognize the objects in the non-discriminative regions. AE-PSL \cite{aepsl} erases the regions with the CAM and forces the network to discover the objects in the remaining regions, but suffers from over-activation, \emph{i.e.}, some background regions are incorrectly activated. GAIN \cite{gain} proposes a two-stage scheme using a shared classifier that minimize the classification score of the masked image using the thresholded CAM generated in the first stage. OC-CSE \cite{occse} improves GAIN by training a segmentation network in the first stage, and using the pre-trained classifier in the second stage. However, Zhang \emph{et al.} \cite{cpn} points out that the AE-based approaches has some weakness. First, their training stage is unstable as some regions might be lost due to the random hiding process. Second, they may over-activate the background regions, resulting in very small classification score in the second stage and thus increase the training instability.

\vspace{-5pt}
\subsection{Vision Transformer}
\vspace{-5pt}

Recently, vision transformers have been introduced to computer vision community. Vision Transformer (ViT) \cite{vit} first proposes to split the input image into a non-overlapping patch sequence. Each patch is then projected into a token using a shared fully-connected (FC) layer. The patch tokens plus a class token will be summed with the corresponding position embedding and passed to the transformer blocks with Multi-Head Self-Attention (MHSA) and Multi-Layer Perceptron (MLP). DeiT \cite{deit} improves ViT via distillation on an extra distillation token. As the MHSA with MLP builds a connection between each token and other tokens, the message passing within the transformer is global and better than the local connection in the conventional CNN. Many approaches aims at improving ViT \cite{pvt,swin,twins,cswin} with better architectures. Recently, TS-CAM \cite{tscam} and MCTFormer \cite{mctformer} introduce transformers to weakly-supervised object localization and WSSS, respectively, and both achieve impressive performance. However, TS-CAM \cite{tscam} and MCTFormer \cite{mctformer} are still classification-based CAM generation approaches and may still suffer from incomplete activation problem. In this paper, we propose a novel transformer-based learning framework for WSSS to generate activation maps under the guidance of semantic, which can produce more complete regions than those CAM-based approaches. 

\vspace{-7pt}
\section{Method}
\vspace{-6pt}

\subsection{Overall Framework}
\vspace{-7pt}

\begin{figure}[!hb]
    \vspace{-10pt}
	\centering
    \includegraphics[width=\textwidth]{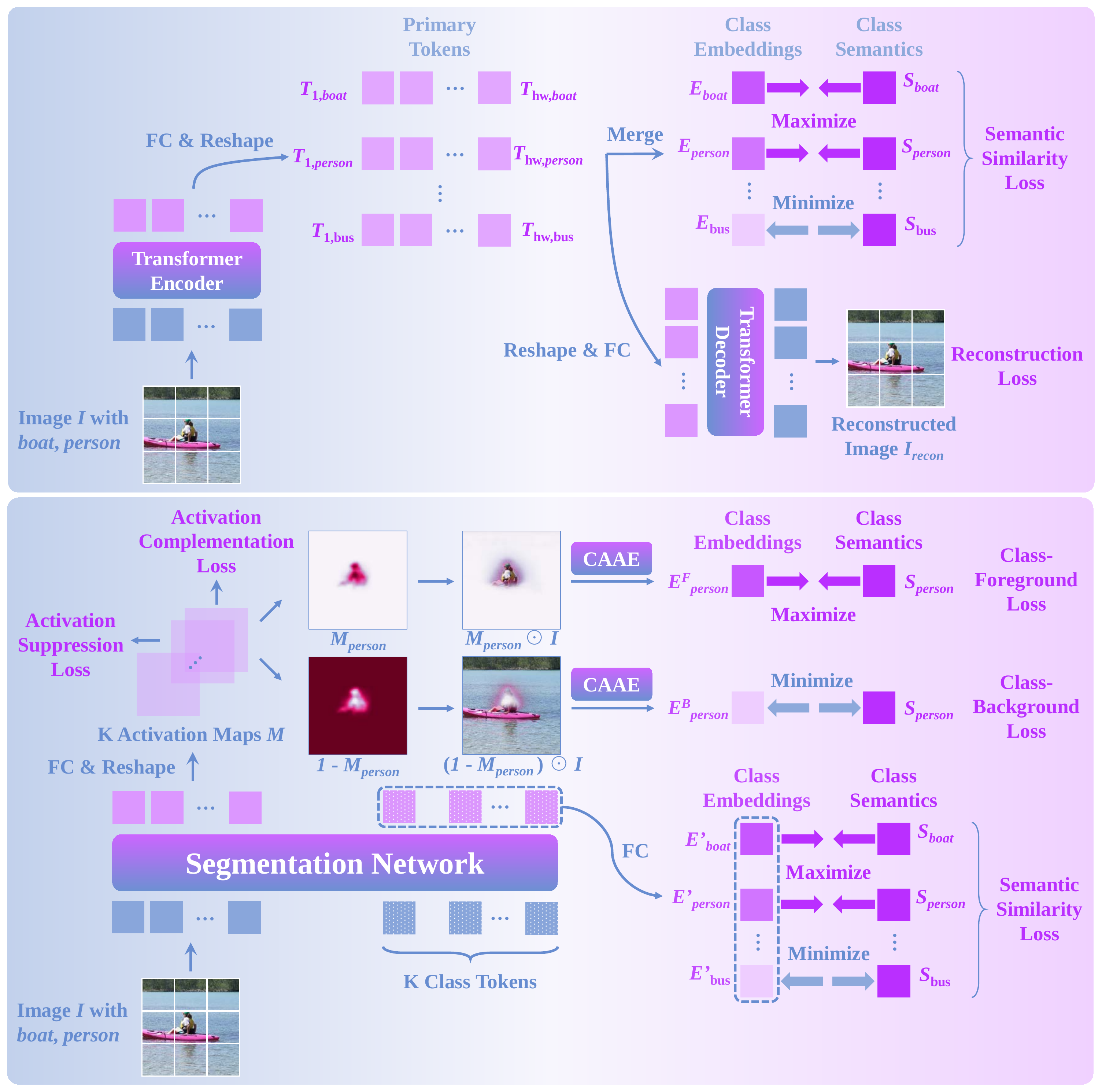}
    \caption{Overview of the proposed two-stage SemFormer. The first stage (top) is to train a class-aware autoencoder (CAAE). The second stage (bottom) is to train the segmentation network using the pre-trained CAAE.}
	\label{fig:overview}
	\vspace{-10pt}
\end{figure}

In this section, we introduce the overall framework of our SemFormer. As shown in Figure \ref{fig:overview}, SemFormer contains two training stages. In the first stage, we train a Class-Aware AutoEncoder (CAAE) under the constraint of semantic similarity and reconstruction loss to correctly extract the class embeddings of the input image, and learn the class semantics of the dataset. In the second stage, we train the segmentation network based on the class embeddings extracted by the pre-trained CAAE and the learned class semantics. The similarity of the same class or different classes are maximized or minimized through four losses (class-foreground, class-background, activation superssion, and activation complementary loss) to guarantee the correctness, completeness, compact, and complementation of activation maps.

\subsection{Class-Aware AutoEncoder}
\label{sec:caae}
\subsubsection{Architecture.}
The Class-Aware AutoEncoder (CAAE) is a transformer-based architecture designed to extract class embeddings for input image and learn class semantics of the dataset, which consists of an encoder (a transformer backbone without classification layer) and a asymmetric decoder (transformer blocks).

\subsubsection{Class-Aware Semantic Representation}
Let $I$ be the input image with $H \times W$ pixels, which is split into $h \times w$ patches, where $h=W/P$, $w=W/P$, and $P$ is the patch size. The patches are flattened and projected to patch tokens $t^0 \in \mathbbm{R}^{hw \times d}$ via a shared FC layer, where $d$ is the dimension of each token embedding. The patch tokens (in our experiments, we find that ignorance of the class token has negligible influence to the performance) are then summed up with the corresponding position embeddings and passed to $L$ transformer blocks with MHSA and MLP. Let $t^L \in \mathbbm{R}^{hw \times d}$ be the patch tokens after the the $L$-th transformer block. $t^L$ is passed to a fully-connected (FC) layer followed by $ReLU$ function and reshaped to generate primary tokens $T \in \mathbbm{R}^{wh \times K \times D}$:
\begin{equation}
\setlength{\abovedisplayskip}{2pt}
\setlength{\belowdisplayskip}{2pt}
\label{eq:crt}
    T = reshape(ReLU(t^LW_1), (wh, K, D)),
\end{equation}
where $W_1 \in \mathbbm{R}^{d \times KD}$ is the weights of the FC layer, $reshape$ denotes the reshaping function, $K$ is the number of classes in the dataset, and $D$ is the class dimension.

After that, we sum up the primary tokens $T$ to generate the class embeddings $E \in \mathbbm{R}^{K \times D}$, \emph{i.e.}, $E = \frac{1}{K} \sum_{i=1}^{wh} T_i$.

The class embeddings $E$ can represent the semantic for each class of the input image. We also define class semantics $S \in \mathbbm{R}^{K \times D}$ to represent the overall semantics of the $K$ classes in the dataset. Each semantic is a $D$-dimension vector corresponds to a class. $S$ is obtained by $S = ReLU(W_S)$, where $W_S$ is the weights, which is randomly initialized before training CAAE and fixed after training.

\subsubsection{Loss Functions}
For CAAE, we design semantic similarity and reconstruction loss to extract the class embeddings of the input image and learn class semantics of the dataset.

\paragraph{Semantic Similarity Loss.}
For the input image $I$, the semantic similarity (SS) loss is designed to make CAAE aware what classes exist in the image. For class $c$, the cosine similarity $Sim(E_c, S_c)$ between the class embedding $E_c$ and the class semantic $S_c$ is maximized if any object of class $c$ exists in the image, otherwise $Sim(E_c, S_c)$ is minimized, where $Sim(a, b) = \frac{a^{\top} b}{\left\|a\right\|_{2}\left\|b\right\|_{2}}$ is the cosine similarity between $a$ and $b$. The value of $Sim(E_c, S_c)$ is in $[0, 1]$ as both $E$ and $S$ are passed to ReLU function without negative values as mentioned above. Let $Y \in \mathbbm{R}^K$ be the image-level annotation of the input image. For class $c$, $Y_c$ is in $\{0, 1\}$, if class $c$ exist in the image, $Y_c=1$, otherwise, $Y_c=0$. The SS loss $\mathcal{L}_{SS}$ is a binary cross entropy (BCE) loss:
\begin{equation}
\setlength{\abovedisplayskip}{0pt}
\setlength{\belowdisplayskip}{0pt}
\label{eq:ss_loss}
    \mathcal{L}_{SS} = -\frac{1}{K}\sum_{c}^{K} [ Y_{c}log(Sim(E_{c}, S_{c})) + (1-Y_{c})log(1 - Sim(E_{c}, S_{c}))].
\end{equation}

\paragraph{Reconstruction Loss.}
There is no guarantee that the embeddings extracted by the encoder can correctly represent the semantic for each class of the input image. To address this issue, we introduce a decoder to perform image reconstruction. The primary tokens $T$ is reshaped and passed to an FC layer to recover the dimension of the patch tokens:
\begin{equation}
\setlength{\abovedisplayskip}{0pt}
\setlength{\belowdisplayskip}{0pt}
\label{eq:crt}
    t'^L = reshape(T, (wh, KD))W_2,
\end{equation}
where $W_2 \in \mathbbm{R}^{KD \times d}$ is the weights of the FC layer, $t'^L$ denotes the recovering result.

The recovering result is then passed to the decoder consisting of transformer blocks followed by an FC layer, and then reshaped to reconstruct the input image. On one hand, as the reconstructed image are generated from the primary tokens $T$, the reconstruction loss can ensure that the primary tokens $T$ have correct semantics of the input image. On the other hand, the embeddings are produced from the primary tokens $T$. As the semantic of $T$ is correct, the semantic of class embeddings can also be ensured. Similar with MAE \cite{mae}, the decoder of CAAE is only used in pre-training CAAE to perform image reconstruction task. Only the encoder is used to extract class embeddings of input image for the training of segmentation network. As MAE \cite{mae} indicates, the design of the decoder architecture can be flexible and independent of the encoder design. In our experiments, we evaluate the impact of the depth (number of transformer blocks) of the decoder.
Let $I_{recon}$ be the reconstructed image from CAAE, the reconstruction loss $\mathcal{L}_{recon}$ is a mean squared error (MSE) loss between $I$ and $I_{recon}$.

\paragraph{Overall Loss.} During training CAAE, the overall loss $\mathcal{L}_{CAAE}$ is defined as:
\begin{equation}
\setlength{\abovedisplayskip}{2pt}
\setlength{\belowdisplayskip}{2pt}
\label{eq:caae_loss}
    \mathcal{L}_{CAAE} = \mathcal{L}_{SS} + \mathcal{L}_{recon}.
\end{equation}

\subsection{Segmentation Network}
\label{sec:seg_net}
\subsubsection{Architecture.}
The segmentation network is a vanilla transformer backbone to be trained for the generation of activation maps with image-level supervision. The patch tokens with $K$ class tokens are summed up with the position embedding and input to the segmentation network. Let $t'^l \in \mathbbm{R}^{K + h'w' \times d'}$ be the tokens of the $l$-th transformer block of the segmentation network, where $d'$ is the dimension of the tokens. The patch tokens $t^{patch}=t'^{L'}_{K + 1:K + h'w', :}$ are then passed to an FC layer follow by the sigmoid function and then transposed and reshaped to generate the activation maps $M$:
\begin{equation}
\setlength{\abovedisplayskip}{2pt}
\setlength{\belowdisplayskip}{2pt}
\label{eq:act_map}
    M = reshape((t^{patch}W^{patch})^{\top}, (K, h', w')),
\end{equation}
where $W^{patch} \in \mathbbm{R}^{d' \times K}$ denotes the weights of the FC layer. Each channel of the activation maps corresponds to a class of the dataset.

\subsubsection{Semantic Guided Activation.}
The CAAE aims to embed the class semantic into the segmentation network to guide the generation of the activation maps. The weights of CAAE are fixed during training of the segmentation network.

\subsubsection{Class-Relative Attention.}
The MHSA in the transformer, which builds token-to-token attentions, can reflect the relationship between the tokens. Hence, the attention from each class token to patch tokens (class-relative attention) can also work as the CAM. Let $\mathcal{A}^l \in \mathbbm{R}^{(K + h'w') \times (K + h'w')}$ be the token-to-token attentions from the $l$-th transformer block (averaged over heads of MHSA), where $h'$ and $w'$ denotes the height and width of the last feature layer of the segmentation network. The class-relative attention of the $l$-th transformer block is $A^l=\mathcal{A}^l_{1:K, K + 1:K + h'w'}$. As the deeper transformer block has stronger semantic, we average the class-relative attentions from the last $U$ (set to 4 in our experiments) transformer blocks and use min-max normalization to generate the final class-relative attention $A'$:
\begin{equation}
\setlength{\abovedisplayskip}{0pt}
\setlength{\belowdisplayskip}{0pt}
\label{eq:cra}
    A' = reshape(min\_max\_norm(\frac{1}{U} \sum_{l=L' - U + 1}^{L'} A^l), (K, h', w')),
\end{equation}
where $L'$ denotes the number of total transformer blocks, $min\_max\_norm$ denotes min-max normalization. 

The final activation map $\mathcal{M}_c = M_c \odot A'_c$ of class $c$ is generated by combining $M_c$ and $A'_c$. In our experiments, we find that the performance of combining the semantic guided activation map with class-relative attention is much better than that of semantic guided activation map. This result indicates that, the learned multi-head self-attention is very useful for semantic segmentation.

\subsubsection{Loss Functions}

\begin{wrapfigure}{!t!br}{0.68\textwidth}
	\vspace{-18pt}
	\begin{center}
		\setlength{\fboxrule}{0pt}
		\fbox{\includegraphics[width=0.66\textwidth]{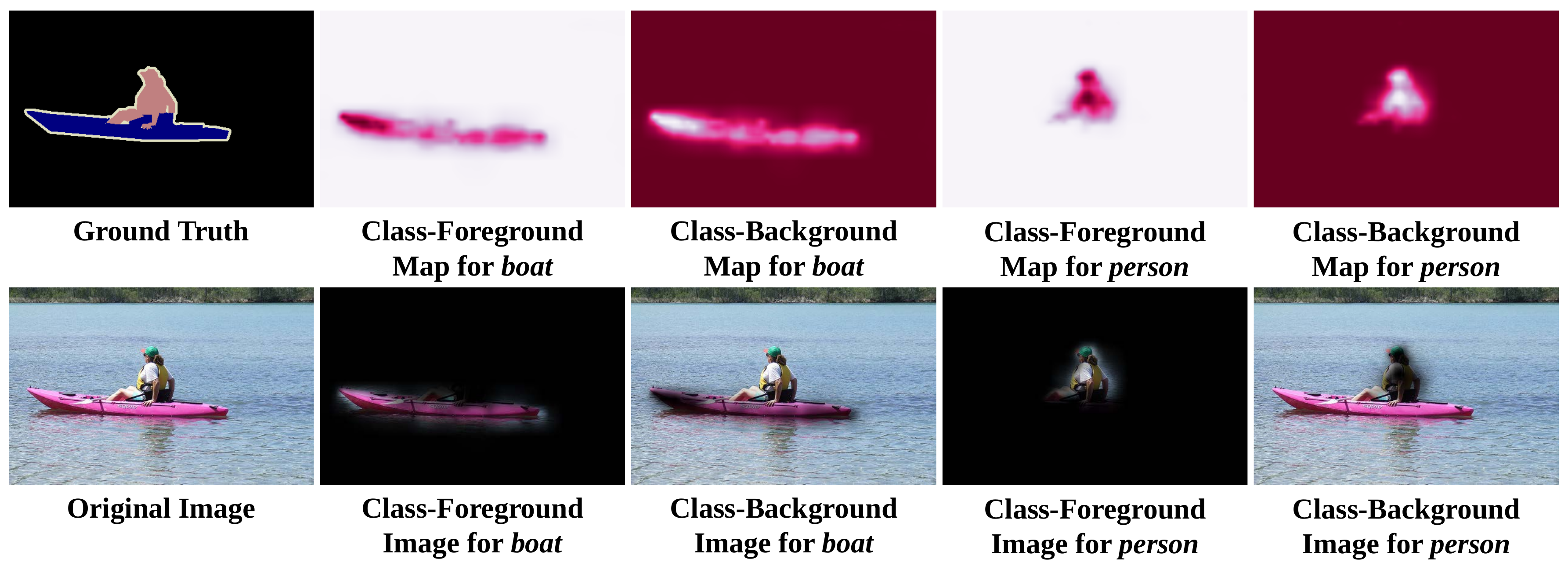}}
	\end{center}	
	\captionsetup{font={scriptsize}}
	\vspace{-12pt}
	\caption{Examples of class-foreground/class-background maps and class-foreground/class-background images generated by the activation maps.}
	\label{fig:act_map}
	\vspace{-10pt}
\end{wrapfigure}

For the segmentation network, we design four losses (class-foreground, class-background, activation suppression, and activation complementation loss) to make the segmentation network aware of the class embeddings.

For a given image $I$, and its CAM $M_c$ of class $c$, a complementary image-pair, \emph{i.e.}, class-foreground image $I^F_c=M'_c \odot I$ and class-background image $I^B_c=(\mathbf{1} - M'_c) \odot I$ are constructed by element-wise multiplication $\odot$, where $M'_c$ is generated by interpolating $M_c$ to the size of input image. We perform class-foreground loss and class-background loss on the class-foreground image and class-background image via the CAAE, to ensure that there is only the semantic of this class in the class-foreground image and no information of this class is available in the class-background image. The adversarial training of the two complementary losses guides the generation of the CAM and ensures its correctness.

As shown in Figure \ref{fig:act_map}, given an input image with \textit{boat} and \textit{person} classes, for class $c \in \{boat, person\}$, the class-foreground image $I_c^F$ is constructed by multiplying the class-foreground map (\emph{i.e.}, activation map) $M_c$ with the input image and used to calculate class-foreground loss, and the class-background image $I_c^B$ is constructed by multiplying the class-background map $\mathbf{1} - M_c$ with the input image for class-background loss. We will give detailed introduction of the four losses in the following sections.

\paragraph{Class-Foreground Loss.} The class-foreground (CF) loss aims at making the segmentation network aware whether the activated regions are correct. For class $c$, the class-foreground images should only contain regions of class $c$, without activating regions of other classes, as shown in Figure \ref{fig:act_map}. Let $E_c^F$ be the embedding for class $c$ of class-foreground image, and $S_c^F$ be the semantic of class $c$. The CF loss is designed to maximize $sim(E_c^F, S_c^F)$, and minimize $Sim(E_{c'}^F, S_{c'}^F)$, where $c' \neq c$. In our experiments, we find that for training the segmentation network, the losses without log function have better results. As a result, the CF loss $\mathcal{L}_{CF}$ is defined as:
\begin{equation}
\setlength{\abovedisplayskip}{2pt}
\setlength{\belowdisplayskip}{2pt}
\label{eq:cf_loss}
    \mathcal{L}_{CF} = \frac{1}{\mathcal{C}} \sum_{c}^{K} \mathbbm{1}(Y_{c}=1)[1 - Sim(E_{c}^{F}, S_{c})] + \frac{1}{\mathcal{C} - 1} \sum_{c', c' \neq c}^{K} \mathbbm{1}(Y_{c}=1, Y_{c'}=1) Sim(E_{c'}^{F}, S_{c'}),
\end{equation}
where $\mathcal{C}$ denotes the number of available classes of the input image, and $\mathbbm{1}(condition) \in \{0, 1\}$ is the indicator function and equal to $1$ only when the $condition$ is true.

\paragraph{Class-Background Loss.} The CF loss encourage to activate the most discriminative regions of the objects, similar with CAM-based approaches. In this situation, some regions of the objects are recognized as background, resulting in incomplete activation. To address this issue, we propose a class-background loss, to minimize the similarity between the embedding of the class-background image $E_c^B$ and the semantic $S_c$ for class $c$, to ensure that there exist no information of this class in the class-background image. Besides, the CB loss also encourages to activate the regions of other available classes in the class-background image to promot the complementation of the class-background image to the class-foreground image, by maximizing the similarity between embedding of other available classes $E_{c'}^B$ and the corresponding semantic $S_{c'}$, where $c' \neq c$. The CB loss $\mathcal{L}_{CB}$ is defined as:
\begin{equation}
\setlength{\abovedisplayskip}{0pt}
\setlength{\belowdisplayskip}{0pt}
\label{eq:cb_loss}
    \mathcal{L}_{CB} = \frac{1}{\mathcal{C}} \sum_{c}^{K} \mathbbm{1}(Y_{c}=1)Sim(E_{c}^{B}, S_{c}) + \frac{1}{\mathcal{C} - 1} \sum_{c', c' \neq c}^{K} \mathbbm{1}(Y_{c}=1, Y_{c'}=1) [1 - Sim(E_{c'}^{B}, S_{c'})],
\end{equation}
where $\mathcal{C}$ denotes the number of available classes of the input image.

The adversarial training between the CF loss and CB loss is similar with the erasing-based methods. However, the erasing-based approaches usually erase the input image using binary mask via hard threshold, which need to be finely tuned and brings difficulty in optimization. Besides, the erasing-based approaches often randomly select a class for training in each training iteration. Zhang \emph{et al.} \cite{cpn} indicates that, the randomness makes the training of the erasing-based approaches unstable. In contrast, we use soft activation, instead of hard-thresholded binary mask, to multiply with the image, and take all available classes into training. As a result, our approach can alleviate the difficulty of optimization due to the hard threshold and instability caused by randomness.

\paragraph{Activation Suppression Loss.} The segmentation network can produce correct and complete activated regions via the adversarial training between CF loss and CB loss. The CB loss will decrease when the activated regions in the class-foreground image enlarges, until they cover the complete object. The CF loss will increase when the activated regions are larger than the object. However, when the object and background look similar, \emph{e.g.}, the spindrift surrounding a white boat in the ocean, the adversarial training between CF and CB loss will converge to a trivial case, \emph{i.e.}, activating large regions of background. We propose an activation suppression (AS) loss to shrink the activated regions in the class-foreground image, which is defined as:
\begin{equation}
\setlength{\abovedisplayskip}{2pt}
\setlength{\belowdisplayskip}{2pt}
\label{eq:as_loss}
    \mathcal{L}_{AS} = \frac{\sigma}{K H W} \sum_c^K \sum_i^H \sum_j^W M_{c}(i, j),
\end{equation}
where $\sigma$ (set to 0.075) is the scaling value of the AS loss.

\paragraph{Activation Complementation Loss.} In our study, we find that some pixels are activated in multiple channels of the class-foreground maps, \emph{e.g.}, some pixels of the train image are activated both in train and background classes, while some pixels of a black regions within a motorbike are not activated in any channels. In these situations, the activation values are not complementary. To address issue, we propose an activation complementation (AC) loss to ensure that each pixel is activated in only one of the classes. We first generate a background map $N_c$ for each class $c$ indicating whether a pixel does not belong to this class, by aggregating class-foreground maps from other classes:
\begin{equation}
\setlength{\abovedisplayskip}{2pt}
\setlength{\belowdisplayskip}{2pt}
\label{eq:background_map}
\begin{split}
    N_{c}(i, j) = max(\{M_{c'}(i, j)\ |\ c' \in \{1, \dots, K\}, c' \neq c\}),
\end{split}
\end{equation}
where $(i, j)$ denotes the pixel coordinates.

The AC loss forces the summation of $M_c$ and $N_c$ to $\mathbf{1}$:
\begin{equation}
\setlength{\abovedisplayskip}{2pt}
\setlength{\belowdisplayskip}{2pt}
\label{eq:ac_loss}
    \mathcal{L}_{AC} = \frac{1}{\mathcal{C} H W} \sum_{c, Y_{c}=1}^{K} \sum_i^H \sum_j^W (M_{c}(i, j) + N_{c}(i, j)\ - 1)^2,
\end{equation}
where $H$ and $W$ are the height and width of the activation map, respectively. With the help of CF and CB loss, the segmentation network will activate the correct class for each pixel.

\paragraph{Semantic Similarity Loss.} The class tokens $t^{class}=t'^{L'}_{1:K + 1, :}$ are used to learn the class-relative attention described in Section \ref{sec:seg_net}, which can improve the segmentation performance. For class $c$, the class token $t^{class}_c$ is passed to a linear layer to produce embedding $E'_{c}$ of this class: $E'_c = t^{class}_cW^{class}_c$,

where $W^{class}_c \in \mathbbm{R}^{d' \times D}$ is the weights of the linear layer. We learn the class-relative attention by conducting semantic similarity loss $\mathcal{L}'_{SS}$ on the class-relative embedding produced by the segmentation network and the corresponding learned class-relative semantic of the dataset:
\begin{equation}
\setlength{\abovedisplayskip}{0pt}
\setlength{\belowdisplayskip}{0pt}
\label{eq:ss_loss_seg}
    \mathcal{L}'_{SS} = -\frac{1}{K}\sum_{c}^{K} [ Y_{c}log(Sim(E'_{c}, S_{c})) + (1-Y_{c})log(1 - Sim(E'_{c}, S_{c}))].
\end{equation}

\paragraph{Overall Loss.} During training the segmentation network, the overall loss is defined as:
\begin{equation}
\setlength{\abovedisplayskip}{2pt}
\setlength{\belowdisplayskip}{2pt}
\label{eq:seg_loss}
    \mathcal{L}_{seg} = \mathcal{L}_{CF} + \mathcal{L}_{CB} + \mathcal{L}_{AS} + \mathcal{L}_{AC} + \mathcal{L}'_{SS}.
\end{equation}

\section{Experiments}
\label{sec:experiments}

\subsection{Experimental Setup}
\label{sec:exp_setup}
\paragraph{Dataset and Evaluation Metric.}
All experiments are conducted on PASCAL VOC 2012 dataset \cite{voc} containing 21 classes (20 object classes and background class). We use the \emph{trainaug} split (10,528 images) with only image-level annotations for training. We use the \emph{train} split (1,464 images) to validate our method. We use the \emph{val} (1,449 images) and \emph{test} (1,456 images) split for evaluating our WSSS approach and comparison with recent state-of-the-art methods, respectively. All the experimental results are reported in the standard mean Intersection over Union (mIoU) metric.

\paragraph{Reproducibility.}
For Class-Aware AutoEncoder (CAAE), the backbone is pre-trained on ImageNet \cite{imagenet}. Stochastic gradient descent (SGD) optimizer with batch size 64 is used to train CAAE for 200 epochs. The poly-policy training strategy is adopted with the initial learning rate of 0.025 and decay power of 0.9. Input images are scaled randomly and then cropped to $224 \times 224$ pixels. For segmentation network, we follow \cite{tscam,mctformer} to use DeiT-S \cite{deit} as backbone for fair comparisons. SGD with batch size 8 is used to train the segmentation network for 20 epochs. The poly-policy training strategy is adopted with the initial learning rate and decay power is 0.005 and 0.9, respectively. Input images are scaled randomly and then cropped to $448 \times 448$ pixels. Experiments are conducted on 2 NVIDIA A100 GPUs.

\subsection{Main Results}

\subsubsection{Performance of CAM and Pseudo Mask}
We first compare the performance of CAM and pseudo mask in Table \ref{tab:cam}. As shown in the 2nd column of the table, our SemFormer achieves \textbf{63.7}\% mIoU, which is 23.4\% higher than TS-CAM \cite{tscam} (We use the official code for re-implementation on PASCAL VOC 2012 dataset.), and outperforms numerous recent state-of-the-art methods, \emph{e.g.}, 2.0\% and 1.6 \% higher than that of MCTFormer \cite{mctformer} and AMN \cite{amn}, respectively. We also compare the performance of the pseudo mask, yielded via AffinityNet \cite{affinitynet}, which is widely adopted by most state-of-the-art methods. As shown in the last column of Table \ref{tab:cam}, the pseudo mask of our SemFormer achieves the best mIoU of \textbf{73.2}\%.

\begin{table}[h]
	\centering
	\begin{minipage}[t]{0.42\linewidth}
	\centering
    \caption{Performances (mIoU (\%)) of CAM and pseudo mask (denoted as Mask) of different methods.}
    \renewcommand\tabcolsep{2pt}
    \begin{tabular}{lcc}
	\toprule
    Method & CAM & Mask \\
	\midrule
	$\text{CAM}_{\text{~~CVPR'16}}$~\cite{cam}                 & 47.4           & -\\
    $\text{SC-CAM}_{\text{~~CVPR'20}}$~\cite{sc-cam}              & 50.9           & 63.4\\
    $\text{SEAM}_{\text{~~CVPR'20}}$~\cite{seam}                  & 55.4           & 63.6\\
    $\text{CONTA}_{\text{~~NeurIPS'20}}$~\cite{conta}             & 56.2           & 67.9\\
    $\text{VWE}_{\text{~~IJCAI'21}}$~\cite{vwe}                   & 55.1           & 67.7\\
    $\text{AdvCAM}_{\text{~~CVPR'21}}$~\cite{advcam}              & 55.6           & 68.0\\
    $\text{TS-CAM}_{\text{~~ICCV'21}}$~\cite{tscam}               & 40.3           & - \\
    $\text{ECS-Net}_{\text{~~ICCV'21}}$~\cite{ecsnet}             & 56.6           & 67.8\\
    $\text{OC-CSE}_{\text{~~ICCV'21}}$~\cite{occse}               & 56.0           & 66.9\\
    $\text{CPN}_{\text{~~ICCV'21}}$~\cite{cpn}                    & 57.4           & 67.8\\
    $\text{RIB}_{\text{~~NeurIPS'21}}$~\cite{rib}                 & 56.5           & 68.6\\
    $\text{CLIMS}_{\text{~~CVPR'22}}$~\cite{clims}                & 56.6           & 70.5\\
    $\text{AMN }_{\text{~~CVPR'22}}$~\cite{amn}                   & 62.1           & 72.2\\
    $\text{MCTFormer}_{\text{~~CVPR'22}}$~\cite{mctformer}        & 61.7           & 69.1\\
    \textbf{SemFormer (ours)}                                & \textbf{63.7}  & \textbf{73.2}\\
	\bottomrule
    \end{tabular}
	\label{tab:cam}
	\vspace{-10pt}
	\end{minipage}\hfill
    \begin{minipage}[t]{0.55\linewidth}
    \centering
    \caption{Performances (mIoU (\%)) of different WSSS methods on PASCAL VOC 2012 \textit{val} and \textit{test} split. \textit{Sup}. denotes the supervision level, including pixel-level ($\mathcal{F}$), box-level ($\mathcal{B}$), saliency-level ($\mathcal{S}$) and image-level ($\mathcal{I}$).}
    \renewcommand\tabcolsep{6pt}
    \begin{tabular}{lccc}
	\toprule
    Method & \textit{Sup}. & \textit{val} & \textit{test}\\
	\midrule
	$\text{DeepLabV1}_{\text{~~ICLR'15}}$~\cite{deeplabv1}                & & 68.7 & 71.6 \\
    $\text{DeepLabV2}_{\text{~~TPAMI'18}}$~\cite{deeplabv2}             & \multirow{-2}{*}{$\mathcal{F}$} & 77.7 & 79.7 \\
	\hline
	$\text{BCM}_{\text{~~CVPR'19}}$~\cite{bcm}                            & & 70.2 & - \\
	$\text{BBAM}_{\text{~~CVPR'21}}$~\cite{bbam}                          & \multirow{-2}{*}{$\mathcal{I} + \mathcal{B}$} & 73.7 & 73.7 \\
	\hline
	$\text{ICD}_{\text{~~CVPR'20}}$~\cite{icd}                            & & 67.8 & 68.0 \\
	$\text{EPS}_{\text{~~CVPR'21}}$~\cite{eps}                            & \multirow{-2}{*}{$\mathcal{I} + \mathcal{S}$} & 71.0 & 71.8 \\
	\hline
	$\text{BES}_{\text{~~ECCV'20}}$~\cite{bes}                            & & 65.7 & 66.6 \\
	$\text{CONTA}_{\text{~~NeurIPS'20}}$~\cite{conta}                     & & 66.1 & 66.7 \\
	$\text{AdvCAM}_{\text{~~CVPR'21}}$~\cite{advcam}                      & & 68.1 & 68.0 \\
	$\text{OC-CSE}_{\text{~~ICCV'21}}$~\cite{occse}                       & & 68.4 & 68.2 \\
	$\text{RIB}_{\text{~~NeurIPS'21}}$~\cite{rib}                         & & 68.3 & 68.6 \\
	CLIMS \footnotesize{CVPR'22 \cite{clims}}                        & & 70.4 & 70.0\\
    $\text{MCTFormer}_{\text{~~CVPR'22}}$~\cite{mctformer}                & & 71.9 & 71.6\\
    \textbf{SemFormer (ours)}                                        & \multirow{-8}{*}{$\mathcal{I}$} & \textbf{74.3} & \textbf{74.0} \\
	\bottomrule
    \end{tabular}
	\label{tab:wsss}
	\end{minipage}
\end{table}

\subsubsection{Performance of WSSS}
We now compare the performance of WSSS with state-of-the-art methods using different supervisions, including image-level labels ($\mathcal{I}$), saliency maps ($\mathcal{S}$), bounding boxes ($\mathcal{B}$), and pixel-level masks ($\mathcal{F}$), in Table \ref{tab:wsss}. We follow \cite{seam,cpn,occse,ecsnet,mctformer} to adopt DeeplabV1 \cite{deeplabv1} with ResNet38 \cite{resnet38} backbone for the fully supervised segmentation network. As shown in the table, our SemFormer achieves \textbf{74.3}\% and \textbf{74.0}\% mIoU on PASCAL VOC 2012 \textit{val} and \textit{test} split, respectively, surpassing many recent state-of-the-art methods, \emph{e.g.}, CLIMS \cite{clims} and MCTFormer \cite{mctformer} by a large margin. Our SemFormer also achieves better performance than some methods using stronger supervisions. For exmaple, the mIoU of our SemFormer is 3.3\% and 2.2\% higher than EPS \cite{eps} using extra saliency maps, on \textit{val} and \textit{test} split, respectively, and surpasses BBAM \cite{bbam} trained with extra bounding-box annotations. These results justify the advantage of our method.

\subsection{Ablation Studies}

\subsubsection{The loss function and class-relative attention of the segmentation network.}

\begin{figure}[h]
	\centering
    \includegraphics[width=\textwidth]{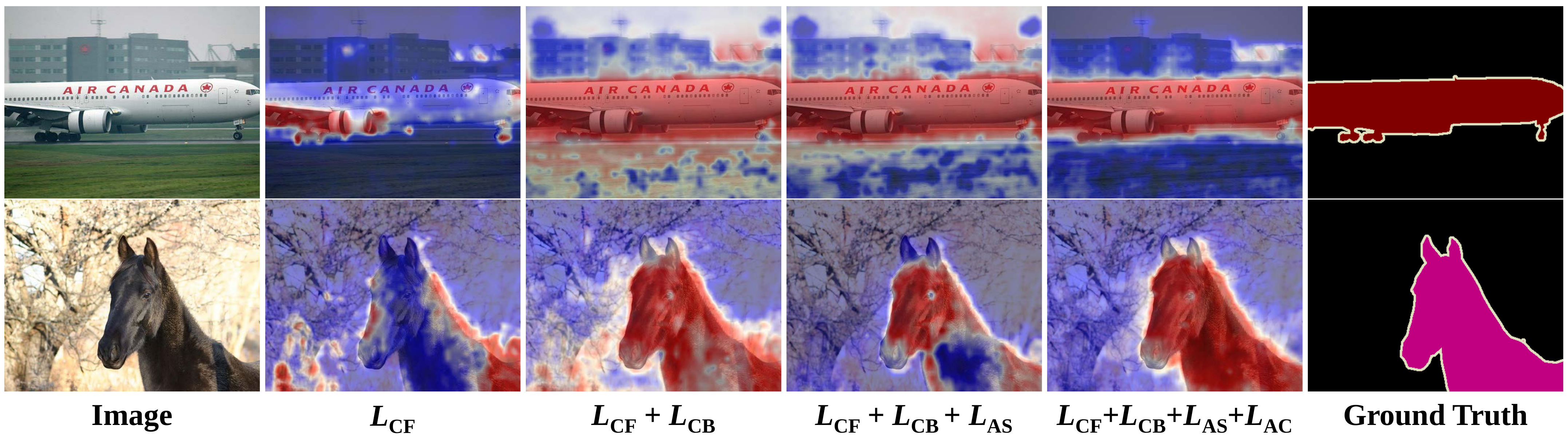}
    \caption{Visualizations of different loss combinations for segmentation network. The 2nd to 5th columns show activation maps generated by different loss combinations.}
	\label{fig:act_map}
\end{figure}

In this section, we evaluate the impact of each loss function of segmentation network. Figure \ref{fig:vis_loss_seg} shows some examples of different loss combinations. We can observe two main problems when only $\mathcal{L}_{CF}$ is used (2nd column), \emph{i.e.}, the activated regions of objects are incomplete, and some background similar with objects are falsely activated. The inclusion of $\mathcal{L}_{CB}$ (3rd column) removes background information in the class-foreground image, and the activated regions are more complete. However, as mentioned above, the adversarial training may activate a large regions in the class-foreground image. The introduction of $\mathcal{L}_{AS}$ (4th column) can address this issue by making the activated regions more compact. The $\mathcal{L}_{AC}$ alleviates the second problem by forcing the segmentation network to activate the correct channel for each pixel with the help of other losses. As shown in the 4th column of the table, the combination of four losses reduces the false activation \emph{e.g.}, the buildings near the aeroplane. Table \ref{tab:seg_loss} lists the mIoUs of different loss combinations. As shown in the table, the combination of all losses achieves the best mIoU of 55.8\%, which is much higher than TS-CAM.

\begin{table}[h]
    \vspace{-8pt}
	\centering
	\begin{minipage}[t]{0.70\linewidth}
	\centering
	\renewcommand\arraystretch{0.85}
    \caption{The impact of each loss function of segmentation network.}
    \renewcommand\tabcolsep{2.5pt}
    \begin{tabular}{c|c|cccccccc}
	\toprule
                        & TS-CAM & \multicolumn{8}{c}{SemFormer (ours)} \\
    \midrule
     $\mathcal{L}_{CF}$ &   -  & \cmk & \cmk & \cmk & \cmk & \cmk & \cmk & \cmk & \cmk\\
     $\mathcal{L}_{CB}$ &   -  &      & \cmk &      &      & \cmk & \cmk &      & \cmk\\
     $\mathcal{L}_{AS}$ &   -  &      &      & \cmk &      & \cmk &      & \cmk & \cmk\\
     $\mathcal{L}_{AC}$ &   -  &      &      &      & \cmk &      & \cmk & \cmk & \cmk\\
     \hline
    mIoU (\%)          & 40.3 & 35.6 & 43.5 & 40.1 & 42.0 & 48.8 & 48.8 & 55.0 &\textbf{55.8}\\
	\bottomrule
    \end{tabular}
	\label{tab:seg_loss}
	\end{minipage}\hfill
	\begin{minipage}[t]{0.27\linewidth}
	\centering
    \caption{The impact of class-relative attention (CRA).}
    \renewcommand\tabcolsep{7pt}
    \begin{tabular}{cc}
	\toprule
	CRA & mIoU (\%) \\
	\midrule
    \xmk & 55.8\\
	\cmk  & \textbf{63.7}\\
	\bottomrule
    \end{tabular}
	\label{tab:cra}
	\end{minipage}
\end{table}

We also evaluate the effectiveness of class-relative attention (CRA) in Table \ref{tab:cra}. As shown in the table, CRA brings dramatic improvement of 7.9\% mIoU to segmentation performance, which justifies the effectiveness of CRA.

\subsubsection{The impact of various token merging of CAAE}
We explore the impact of various token merging of CAAE to the segmentation performance in Table \ref{tab:tok_merge}. As shown in the table, during training of CAAE, using ``sum'' reduction to merge the primary tokens into the class embedding is around 2\% (mIoU) better than ``mean'' or ``max'' reduction. This can be explained by the representation quality of the embedding. ``max'' reduction only focuses on the most discriminative features and ignores the non-discriminative ones. As ``mean'' reduction normalizes the embeddings with a factor, \emph{i.e.}, number of primary tokens, the grads propagated from the embeddings to the primary tokens are also reduced by the factor, which may influence the learning of class semantic. ``sum'' reduction takes all tokens into consideration and avoid influencing the learning of class semantic. The above results indicate that the summation of the primary tokens may be more properiate to represent the class semantic than mean or maximum.

\begin{table}[!t!bh]
	\centering
	\begin{minipage}[t]{0.21\linewidth}
	\centering
	\renewcommand\arraystretch{0.85}
    \caption{The impact of various token merging (Mer.) of CAAE.}
    \renewcommand\tabcolsep{2.8pt}
    \begin{tabular}{lc}
	\toprule
    Mer. & mIoU (\%) \\
	\midrule
	sum & \textbf{63.7}\\
	mean & 61.6\\
	max & 61.8\\
	\bottomrule
    \end{tabular}
	\label{tab:tok_merge}
	\end{minipage}\hfill
	\begin{minipage}[t]{0.22\linewidth}
	\centering
	\renewcommand\arraystretch{0.85}
    \caption{The impact of different encoder backbone (Bac.) of CAAE.}
    \renewcommand\tabcolsep{2pt}
    \begin{tabular}{lc}
	\toprule
    Bac. & mIoU (\%) \\
	\midrule
	DeiT-S & 59.6\\
	DeiT-B & 61.3\\
	DeiT-B\alambic & \textbf{63.7}\\
	\bottomrule
    \end{tabular}
	\label{tab:enc_bac}
	\end{minipage}\hfill
    \begin{minipage}[t]{0.22\linewidth}
    \centering
    \renewcommand\arraystretch{0.88}
    \caption{The impact of decoder depth of CAAE.}
    \renewcommand\tabcolsep{3pt}
    \begin{tabular}{lc}
	\toprule
    Blocks & mIoU (\%)\\
	\midrule
	2 & 61.5\\
	4 & 63.2\\
	8 & \textbf{63.7}\\
	10 & 61.6\\
	\bottomrule
    \end{tabular}
	\label{tab:dec_dep}
	\end{minipage}\hfill
	\begin{minipage}[t]{0.21\linewidth}
	\centering
	\renewcommand\arraystretch{0.88}
    \caption{The impact of dimension (Dim.) of CRE.}
    \renewcommand\tabcolsep{2pt}
    \begin{tabular}{lc}
	\toprule
    Dim. & mIoU (\%)\\
	\midrule
	32  & 58.8 \\
	128 & 61.7\\
	256 & \textbf{63.7}\\
	512 & 61.7\\
	\bottomrule
    \end{tabular}
	\label{tab:cls_dim}
	\end{minipage}
    \vspace{-10pt}
\end{table}

\subsubsection{The impact of different encoder backbone of CAAE}
We explore the impact of different encoder backbone of CAAE to the segmentation performance in Table \ref{tab:enc_bac}. As shown in Table \ref{tab:enc_bac}, when the backbone becomes larger (from DeiT-S to DeiT-B\alambic), the segmentation performance becomes better. Taking DeiT-B\alambic as the encoder backbone will bring around 4\% mIoU improvement to the segmentation performance than DeiT-S. This result indicates that the capacity of the encoder of CAAE is highly important to the segmentation result. As the class embeddings generated by the CAAE are important to train the segmentation network, an encoder with higher capacity will produce embeddings with better representation, and thus brings superior segmentation results. Therefore, adopting an encoder with higher capacity of the CAAE is beneficial to the segmentation, without bringing any inference cost to the segmentation network.

\subsubsection{The impact of decoder depth of CAAE}
In this section, we evaluate the impact of the decoder depth (number of transformer blocks) of CAAE in Table \ref{tab:dec_dep}. As shown in Table \ref{tab:dec_dep}, a sufficient depth of the decoder is important to the segmentation performance. This can be explained by the relationship between reconstruction task and the CAAE. The last several layers of the decoders in CAAE are more specialized to the reconstruction task. A reasonable deep decoder with sufficient reconstruction capacity, is desirable for primary tokens to extract more abstract class semantics, which can bring around 2\% improvement in mIoU.

\subsubsection{The impact of dimension for class embedding}
We evaluate the impact of the dimension of the class embedding to the segmentation network, in Table \ref{tab:cls_dim}. As shown in the table, the dimension of 256 achieves the best segmentation performance. We also find that if we decrease or increase the dimension, the segmentation performance will drop. In our opinion, a small dimension is not able to fully represent the semantic of the input image, while the larger dimension may bring redundant features and disturb the optimization.

\subsection{Model Complexity}

\begin{wraptable}{h}{0.52\textwidth}
    \vspace{-12pt}
	\centering
    \caption{Model complexity under $448 \times 448$ input resolution.}
    \renewcommand\tabcolsep{2pt}
    \begin{tabular}{ccc}
    \toprule
    \textbf{Model}   & \textbf{Parameters (M)} & \textbf{FLOPs (G)} \\
    \midrule
    TSCAM            & 22.04                   & 22.67              \\
    SemFormer (ours) & 24.06                   & 23.34             \\
    \bottomrule
    \end{tabular}
    \label{tab:model_complexity}
\end{wraptable}

In this section, we compare the model complexity of the proposed SemFormer with that of the baseline TSCAM \cite{tscam}, in Table \ref{tab:model_complexity}. As shown in the table, compared to the baseline TSCAM, our method have negligible increase of number of parameters and FLOPs, which indicates the high efficiency of our model.

\section{Conclusions}
In this paper, we propose a novel framework, Semantic Guided Activation Transformer (SemFormer), for weakly-supervised semantic segmentation. We design a transformer-based Class-Aware AutoEncoder to learn the semantic for each class and extract class embeddings for input image. The class embeddings of foreground and background images and learned semantics are used to guide the generation of activation maps, with four losses (class-foreground, class-background, activation suppression, and activation complementation loss). Besides, we use the attention learned by transformer to enhance the activation maps. Experiments show that SemFormer achieves considerable performance and surpasses many recent mainstream approaches on PASCAL VOC 2012 dataset.

\clearpage

\appendix

\section{Appendix}

\subsection{More Ablation Studies}

\begin{table}[h]
	\centering
	\begin{minipage}[t]{0.22\linewidth}
	\centering
    \caption{The impact of decoder width of CAAE.}
    \renewcommand\tabcolsep{3.5pt}
    \begin{tabular}{lc}
	\toprule
    Width & mIoU (\%)\\
	\midrule
	384 & 61.5\\
	576 & 62.1\\
	768 & \textbf{63.7}\\
	1152 & 61.0\\
	\bottomrule
    \end{tabular}
	\label{tab:dec_wid}
	\end{minipage}\hfill
	\begin{minipage}[t]{0.22\linewidth}
	\centering
    \caption{The impact of reconstruction loss ($\mathcal{L}_{recon}$) of CAAE.}
    \renewcommand\tabcolsep{3pt}
    \begin{tabular}{cc}
	\toprule
    $\mathcal{L}_{recon}$ & mIoU (\%)\\
	\midrule
	 \xmk & 61.7 \\
	 \cmk  & \textbf{63.7}\\
	\bottomrule
    \end{tabular}
	\label{tab:recon_loss}
	\end{minipage}\hfill
	\begin{minipage}[t]{0.45\linewidth}
	\centering
    \caption{The impact of class-relative attention (CRA) of segmentation network. FP: false positive. FN: false negative.}
    \renewcommand\tabcolsep{3.5pt}
    \begin{tabular}{cccc}
	\toprule
	CRA & mIoU (\%)$\uparrow$ & FP (\%)$\downarrow$ & FN (\%)$\downarrow$\\
	\midrule
    \xmk &    55.8       & 20.5          & 23.7\\
	\cmk  & \textbf{63.7} & \textbf{19.3} & \textbf{17.0}\\
	\bottomrule
    \end{tabular}
	\label{tab:cra}
	\end{minipage}
\end{table}

\subsubsection{The impact of decoder width of CAAE}
In this section, we evaluate the impact of the decoder width (dimension of transformer blocks) of CAAE in Table \ref{tab:dec_wid}. We add a linear layer before the decoder to change the dimension of the tokens. As shown in the table, keeping the same decoder width as the encoder backbone (\emph{i.e.}, 768 for DeiT-B\alambic) achieves the best mIoU of \textbf{63.7}\%. A narrow decoder, \emph{e.g.}, with dimension of 384 can also work well.

\subsubsection{The impact of reconstruction loss of CAAE}
In this section, we evaluate the impact of the reconstruction loss of CAAE in Table \ref{tab:recon_loss}. As shown in the table, the reconstruction loss can bring 2\% mIoU improvement to the segmentation performance, which indicates that the reconstruction loss can ensure the correctness of the semantic of the class embeddings.

\subsubsection{Further analysis of class-relative attention of segmentation network.}
\begin{figure}[h]
	\centering
    \includegraphics[width=0.98\textwidth]{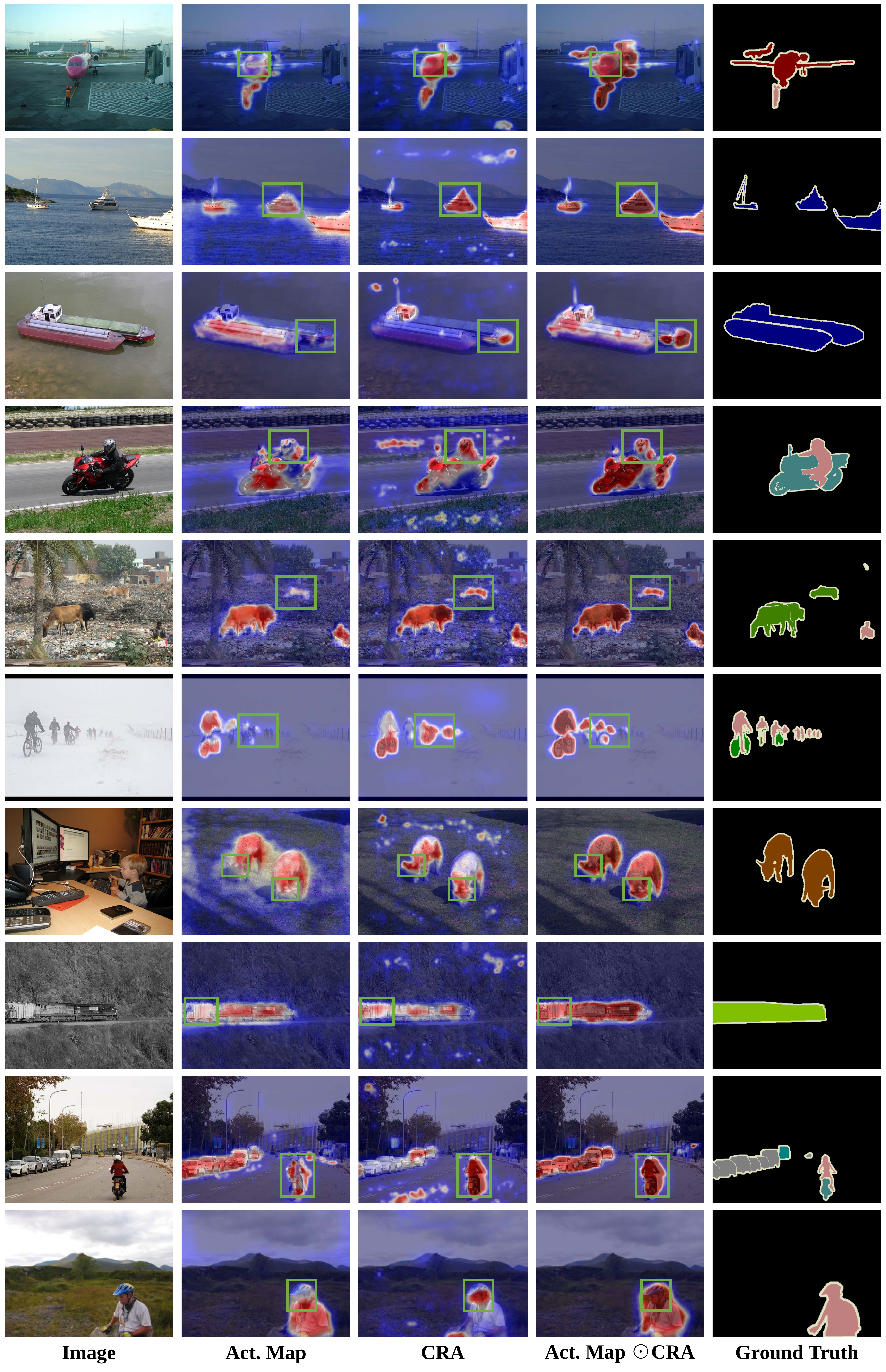}
    \caption{Visualization results of activation map and CRA on PASCAL VOC 2012 \textit{train} split. From left ot right: original image (Image), activation map (Act. Map), class-relative attention (CRA), final activation map (Act. Map $\otimes$ CRA), and Ground Truth. The green boxes show the improvement brought by the CRA to the Act. Map.}
	\label{fig:vis_cra}
	\vspace{-10pt}
\end{figure}

We have roughly evaluated the impact of the class-relative attention (CRA) in the main body. In this section, we further analyze the reason why the CRA can bring considerable improvement to the segmentation performance. We first provide some examples of activation map and CRA in Figure \ref{fig:vis_cra}. As shown in the figure, the CRA is able to activate the missing regions in activation maps (within green boxes), therefore, the CRA can reduce the under-activation (false negative). We also provide a quantitative result of segmentation performance listed in Table \ref{tab:cra}. The CRA can significantly reduce the false negative (FN) by 6.7\%, and also decrease the false positive (FP) by around 1\%, and thus dramatically boosts the segmentation performance.

\subsubsection{Investigation of the number of the last transformer blocks for the CRA}
\begin{table}[h]
\centering
    \caption{Investigation of the number of the last transformer blocks for the CRA.}
    \renewcommand\tabcolsep{6pt}
    \begin{tabular}{cccccccc}
	\toprule
	$U$ & 1 & 2 & 3 & 4 & 5 & 6 & 7\\
	\midrule
	mIoU (\%)$\uparrow$ & 60.0 & 61.6 & 63.3 & \textbf{63.7} & 63.5 & 63.2 & 63.0\\
	FP (\%)$\downarrow$ & 24.0 & 21.8 & 20.0 & \textbf{19.3} & 20.0 & 21.0 & 20.5\\
	FN (\%)$\downarrow$ & 16.0 & 16.6 & 16.6 & 17.0 & 16.5 & \textbf{15.7} & 16.5\\
	\bottomrule
    \end{tabular}
    \label{tab:cra_num}
\end{table}

As described in Section 3.1.2 of the main body, we average the CRA from the last $U$ transformer blocks as the final CRA. We investigate the impact of various $U$s to the segmentation performance in Table \ref{tab:cra_num}. As shown in the table, when $U$ is set to 4, the segmentation network achieves the best mIoU of \textbf{63.7}\%. We also find that the FP drops significantly (from 24.0\% to 19.3\%) when $U$ is changed from 1 to 4. This results indicates that the last $U$ transformer blocks can produce semantically strong CRAs that can alleviate the over-activation problem, and thus improve the segmentation performance.

\subsubsection{Investigation of the effectiveness of image reconstruction module in CAAE}

\begin{wraptable}{h}{0.5\textwidth}
\centering
\caption{Investigation of the effectiveness of image reconstruction module in CAAE.}
\renewcommand\tabcolsep{6pt}
\begin{tabular}{cc}
\toprule
\textbf{image reconstructing module} & \textbf{mIoU (\%)} \\
\midrule
\xmk                                    & 61.7               \\
\cmk                                    & \textbf{63.7}     \\
\bottomrule
\end{tabular}
\label{tab:img_recon}
\end{wraptable}

In this section, we investigate the effectiveness of image reconstruction module in CAAE. Table \ref{tab:img_recon} lists the segmentation performance of without or with the image reconstruction module. The results show that the image reconstruction module is important for segmentation, as the image reconstruction can help to preserve the details of the image in the first CAAE stage, and avoid only extracting the most semantically strong features.

\subsubsection{Investigation of the impact of the refinement from class tokens}
In this section, we investigate the impact of class token refinement. The class tokens are not only used to refine the activation map, but also used to facilitate the segmentation network to learn class semantics through the semantic similarity loss. The segmentation network can benefit from the mutual learning between class tokens and patch tokens, through multi-head self-attention.

\begin{table}[h]
\centering
\caption{Investigation of the impact of the refinement from class tokens.}
\begin{tabular}{ccc}
\toprule
\textbf{Semantic Similarity Loss} & \textbf{Class-Relative Attention} & \textbf{mIoU (\%)} \\
\midrule
                                  &                                 & 54.8               \\
\cmk                              &                                 & 55.8               \\
\cmk                              & \cmk                            & 63.7 \\
\bottomrule
\end{tabular}
\label{tab:token_refine}
\end{table}

Table \ref{tab:token_refine} lists the results of whether using semantic similarity loss or class-relative attention. The 2nd and 3rd rows show that the segmentation network can obtain 1.0\% mIoU improvement from the semantic similarity loss even without class-relative attention, which justifies the effectiveness of the mutual learning between class tokens and patch tokens.

\subsubsection{Investigation of the final activation map}
\begin{wraptable}{h}{0.55\textwidth}
\centering
\caption{Investigation of different final activation map.}
\renewcommand\tabcolsep{3pt}
\begin{tabular}{ccc}
\toprule
\textbf{activatio map} & \textbf{class-relative attention} & \textbf{mIoU (\%)} \\
\midrule
\cmk                   &                                   & 55.8               \\
                       & \cmk                              & 38.3               \\
\cmk                   & \cmk                              & \textbf{63.7}      \\
\bottomrule
\end{tabular}
\label{tab:final_act_map}
\end{wraptable}

In this section, we investigate the performance of different final activation map in Table \ref{tab:final_act_map}. As shown in the table, the performance of using only attention from class token to patch token as the final activation map is somewhat lower than the original result. The attention from class token to patch token may work as a complement to the activation map for refinement.

\subsubsection{Ablation study of the scaling value of the AS loss}
\begin{wraptable}{h}{0.46\textwidth}
\vspace{-12pt}
\centering
\caption{Ablation study of the scaling value of the AS loss.}
\begin{tabular}{ccccc}
\toprule
\textit{$\sigma$} & 0.025 & 0.05 & 0.075 & 0.1 \\
\midrule
mIoU (\%)  & 62.3           & 62.8          & \textbf{63.7}  & 62.8     \\
\bottomrule
\end{tabular}
\label{tab:sigma}
\end{wraptable}

In this section, we provide the ablation study of the scaling value $\sigma$ in Eq. \ref{eq:as_loss} of the main body, in Table \ref{tab:sigma}. As shown in the table, the segmentation network achieves the best mIoU of 63.7\% when $\sigma$ is set to 0.075.

~

\subsection{Visualization Results of WSSS}
\begin{figure}[h]
	\centering
    \includegraphics[width=0.98\textwidth]{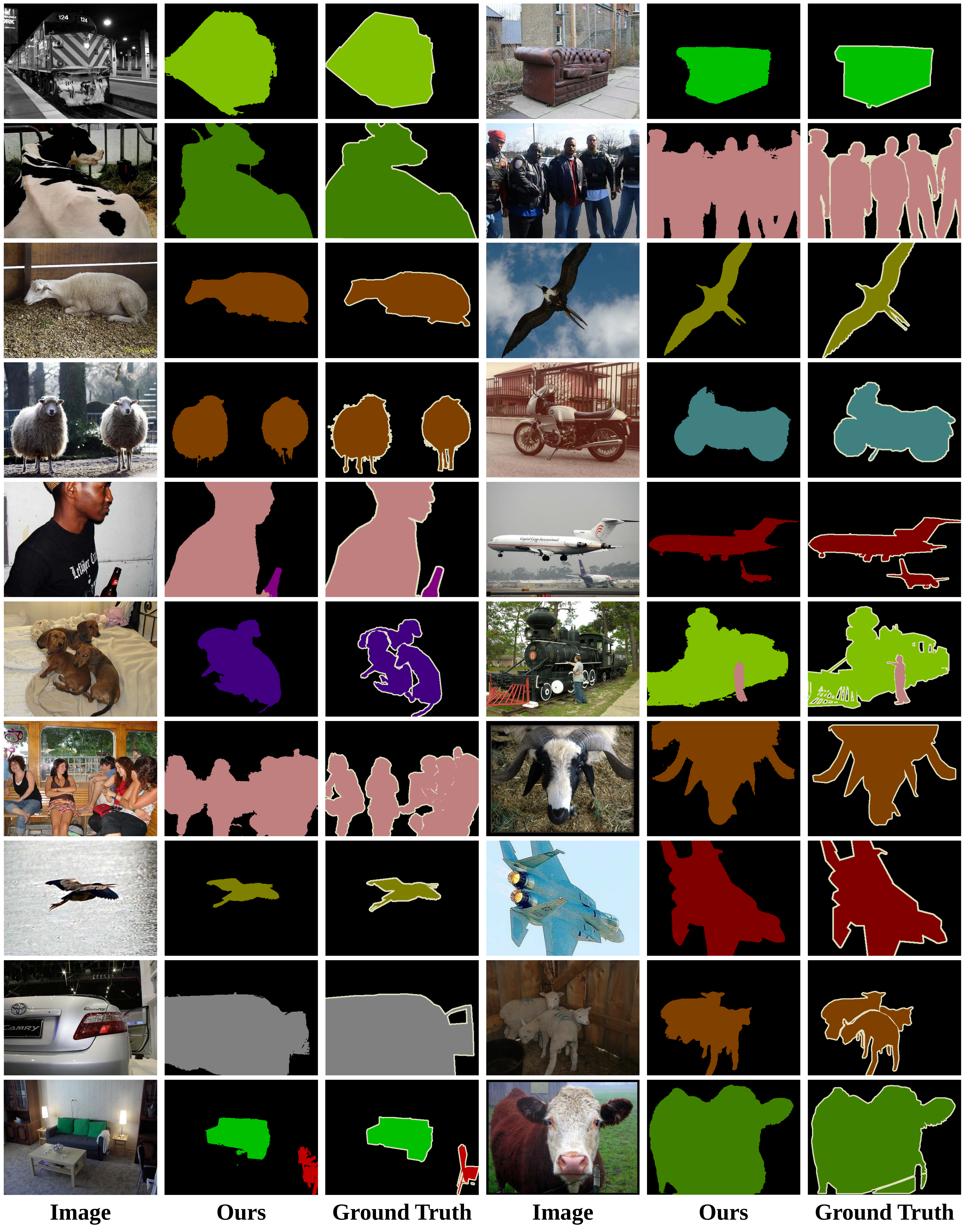}
    \caption{Visualization results of WSSS on PASCAL VOC 2012 \textit{val} split.}
	\label{fig:vis_wsss}
	\vspace{-10pt}
\end{figure}

In this section, we provide visualization results of WSSS in Figure \ref{fig:vis_wsss}.

\clearpage

{
\small
\bibliographystyle{plain}
\bibliography{references}

\begin{thebibliography}{10}

\bibitem{affinitynet}
Jiwoon Ahn and Suha Kwak.
\newblock Learning pixel-level semantic affinity with image-level supervision
  for weakly supervised semantic segmentation.
\newblock In {\em CVPR}, pages 4981--4990, 2018.

\bibitem{whatpoint}
Amy~L. Bearman, Olga Russakovsky, Vittorio Ferrari, and Fei{-}Fei Li.
\newblock What's the point: Semantic segmentation with point supervision.
\newblock In {\em ECCV}, pages 549--565, 2016.

\bibitem{sc-cam}
Yu{-}Ting Chang, Qiaosong Wang, Wei{-}Chih Hung, Robinson Piramuthu, Yi{-}Hsuan
  ai, and Ming{-}Hsuan Yang.
\newblock Weakly-supervised semantic segmentation via sub-category exploration.
\newblock In {\em CVPR}, pages 8988--8997, 2020.

\bibitem{deeplabv1}
{Liang-Chieh} Chen, George Papandreou, Iasonas Kokkinos, Kevin Murphy, and
  Alan~L Yuille.
\newblock Semantic image segmentation with deep convolutional nets and fully
  connected {CRFs}.
\newblock {\em ICLR}, 2015.

\bibitem{bes}
Liyi Chen, Weiwei Wu, Chenchen Fu, Xiao Han, and Yuntao Zhang.
\newblock Weakly supervised semantic segmentation with boundary exploration.
\newblock In {\em ECCV}, pages 347--362, 2020.

\bibitem{deeplabv2}
{Chen, Liang-Chieh}, George Papandreou, Iasonas Kokkinos, Kevin Murphy, and
  Alan~L Yuille.
\newblock {DeepLab}: Semantic image segmentation with deep convolutional nets,
  atrous convolution, and fully connected {CRFs}.
\newblock {\em TPAMI}, 40(4):834--848, 2018.

\bibitem{twins}
Xiangxiang Chu, Zhi Tian, Yuqing Wang, Bo~Zhang, Haibing Ren, Xiaolin Wei,
  Huaxia Xia, and Chunhua Shen.
\newblock Twins: Revisiting the design of spatial attention in vision
  transformers.
\newblock In Marc'Aurelio Ranzato, Alina Beygelzimer, Yann~N. Dauphin, Percy
  Liang, and Jennifer~Wortman Vaughan, editors, {\em NeurIPS}, pages
  9355--9366, 2021.

\bibitem{imagenet}
J.~{Deng}, W.~{Dong}, R.~{Socher}, L.~{Li}, {Kai Li}, and {Li Fei-Fei}.
\newblock Imagenet: A large-scale hierarchical image database.
\newblock In {\em CVPR}, pages 248--255, 2009.

\bibitem{cswin}
Xiaoyi Dong, Jianmin Bao, Dongdong Chen, Weiming Zhang, Nenghai Yu, Lu~Yuan,
  Dong Chen, and Baining Guo.
\newblock Cswin transformer: A general vision transformer backbone with
  cross-shaped windows.
\newblock {\em arXiv preprint arXiv:2107.00652}, 2021.

\bibitem{vit}
Alexey Dosovitskiy, Lucas Beyer, Alexander Kolesnikov, Dirk Weissenborn,
  Xiaohua Zhai, Thomas Unterthiner, Mostafa Dehghani, Matthias Minderer, Georg
  Heigold, Sylvain Gelly, Jakob Uszkoreit, and Neil Houlsby.
\newblock An image is worth 16x16 words: Transformers for image recognition at
  scale.
\newblock In {\em ICLR}, 2021.

\bibitem{voc}
Mark Everingham, SM~Ali Eslami, Luc Van~Gool, Christopher~KI Williams, John
  Winn, and Andrew Zisserman.
\newblock The pascal visual object classes challenge: A retrospective.
\newblock {\em IJCV}, 111(1):98--136, 2015.

\bibitem{icd}
Junsong Fan, Zhaoxiang Zhang, Chunfeng Song, and Tieniu Tan.
\newblock Learning integral objects with intra-class discriminator for
  weakly-supervised semantic segmentation.
\newblock In {\em CVPR}, pages 4282--4291, 2020.

\bibitem{cian}
Junsong Fan, Zhaoxiang Zhang, Tieniu Tan, Chunfeng Song, and Jun Xiao.
\newblock Cian: Cross-image affinity net for weakly supervised semantic
  segmentation.
\newblock In {\em AAAI}, pages 10762--10769, 2020.

\bibitem{tscam}
Wei Gao, Fang Wan, Xingjia Pan, Zhiliang Peng, Qi~Tian, Zhenjun Han, Bolei
  Zhou, and Qixiang Ye.
\newblock {TS-CAM:} token semantic coupled attention map for weakly supervised
  object localization.
\newblock In {\em ICCV}, pages 2866--2875, 2021.

\bibitem{mae}
Kaiming He, Xinlei Chen, Saining Xie, Yanghao Li, Piotr Doll{\'a}r, and Ross
  Girshick.
\newblock Masked autoencoders are scalable vision learners.
\newblock In {\em CVPR}, 2022.

\bibitem{dsrg}
Zilong Huang, Xinggang Wang, Jiasi Wang, Wenyu Liu, and Jingdong Wang.
\newblock Weakly-supervised semantic segmentation network with deep seeded
  region growing.
\newblock In {\em CVPR}, pages 7014--7023, 2018.

\bibitem{oaa}
Peng-Tao Jiang, Qibin Hou, Yang Cao, Ming-Ming Cheng, Yunchao Wei, and Hong-Kai
  Xiong.
\newblock Integral object mining via online attention accumulation.
\newblock In {\em ICCV}, pages 2070--2079, 2019.

\bibitem{sdi}
Anna Khoreva, Rodrigo Benenson, Jan~Hendrik Hosang, Matthias Hein, and Bernt
  Schiele.
\newblock Simple does it: Weakly supervised instance and semantic segmentation.
\newblock In {\em CVPR}, pages 1665--1674, 2017.

\bibitem{sec}
Alexander Kolesnikov and Christoph~H. Lampert.
\newblock Seed, expand and constrain: Three principles for weakly-supervised
  image segmentation.
\newblock In {\em ECCV}, pages 695--711, 2016.

\bibitem{occse}
Hyeokjun Kweon, Sung-Hoon Yoon, Hyeonseong Kim, Daehee Park, and Kuk-Jin Yoon.
\newblock Unlocking the potential of ordinary classifier: Class-specific
  adversarial erasing framework for weakly supervised semantic segmentation.
\newblock In {\em ICCV}, pages 6994--7003, 2021.

\bibitem{rib}
Jungbeom Lee, Jooyoung Choi, Jisoo Mok, and Sungroh Yoon.
\newblock Reducing information bottleneck for weakly supervised semantic
  segmentation.
\newblock In Marc'Aurelio Ranzato, Alina Beygelzimer, Yann~N. Dauphin, Percy
  Liang, and Jennifer~Wortman Vaughan, editors, {\em NeurIPS}, pages
  27408--27421, 2021.

\bibitem{advcam}
Jungbeom Lee, Eunji Kim, and Sungroh Yoon.
\newblock Anti-adversarially manipulated attributions for weakly and
  semi-supervised semantic segmentation.
\newblock In {\em CVPR}, pages 4071--4080, 2021.

\bibitem{bbam}
Jungbeom Lee, Jihun Yi, Chaehun Shin, and Sungroh Yoon.
\newblock {BBAM:} bounding box attribution map for weakly supervised semantic
  and instance segmentation.
\newblock In {\em CVPR}, pages 2643--2652, 2021.

\bibitem{amn}
Minhyun Lee, Dongseob Kim, and Hyunjung Shim.
\newblock Threshold matters in {WSSS:} manipulating the activation for the
  robust and accurate segmentation model against thresholds.
\newblock In {\em CVPR}, 2022.

\bibitem{eps}
Seungho Lee, Minhyun Lee, Jongwuk Lee, and Hyunjung Shim.
\newblock Railroad is not a train: Saliency as pseudo-pixel supervision for
  weakly supervised semantic segmentation.
\newblock In {\em CVPR}, pages 5495--5505, 2021.

\bibitem{gain}
Kunpeng Li, Ziyan Wu, Kuan{-}Chuan Peng, Jan Ernst, and Yun Fu.
\newblock Tell me where to look: Guided attention inference network.
\newblock In {\em CVPR}, 2018.

\bibitem{scribblesup}
Di~Lin, Jifeng Dai, Jiaya Jia, Kaiming He, and Jian Sun.
\newblock Scribblesup: Scribble-supervised convolutional networks for semantic
  segmentation.
\newblock In {\em CVPR}, pages 3159--3167, 2016.

\bibitem{swin}
Ze~Liu, Yutong Lin, Yue Cao, Han Hu, Yixuan Wei, Zheng Zhang, Stephen Lin, and
  Baining Guo.
\newblock Swin transformer: Hierarchical vision transformer using shifted
  windows.
\newblock In {\em ICCV}, pages 9992--10002, 2021.

\bibitem{fcn}
Jonathan Long, Evan Shelhamer, and Trevor Darrell.
\newblock Fully convolutional networks for semantic segmentation.
\newblock In {\em CVPR}, pages 3431--3440, 2015.

\bibitem{vwe}
Lixiang Ru, Bo~Du, and Chen Wu.
\newblock Learning visual words for weakly-supervised semantic segmentation.
\newblock In {\em IJCAI}, 2021.

\bibitem{bcm}
Chunfeng Song, Yan Huang, Wanli Ouyang, and Liang Wang.
\newblock Box-driven class-wise region masking and filling rate guided loss for
  weakly supervised semantic segmentation.
\newblock In {\em CVPR}, pages 3136--3145, 2019.

\bibitem{ecsnet}
Kunyang Sun, Haoqing Shi, Zhengming Zhang, and Yongming Huang.
\newblock Ecs-net: Improving weakly supervised semantic segmentation by using
  connections between class activation maps.
\newblock In {\em ICCV}, pages 7283--7292, 2021.

\bibitem{deit}
Hugo Touvron, Matthieu Cord, Matthijs Douze, Francisco Massa, Alexandre
  Sablayrolles, and Herv{\'{e}} J{\'{e}}gou.
\newblock Training data-efficient image transformers {\&} distillation through
  attention.
\newblock In {\em ICML}, pages 10347--10357, 2021.

\bibitem{rawks}
Paul Vernaza and Manmohan Chandraker.
\newblock Learning random-walk label propagation for weakly-supervised semantic
  segmentation.
\newblock In {\em CVPR}, 2017.

\bibitem{pvt}
Wenhai Wang, Enze Xie, Xiang Li, Deng{-}Ping Fan, Kaitao Song, Ding Liang, Tong
  Lu, Ping Luo, and Ling Shao.
\newblock Pyramid vision transformer: {A} versatile backbone for dense
  prediction without convolutions.
\newblock In {\em ICCV}, pages 548--558, 2021.

\bibitem{seam}
Yude Wang, Jie Zhang, Meina Kan, Shiguang Shan, and Xilin Chen.
\newblock Self-supervised equivariant attention mechanism for weakly supervised
  semantic segmentation.
\newblock In {\em CVPR}, pages 12275--12284, 2020.

\bibitem{aepsl}
Yunchao Wei, Jiashi Feng, Xiaodan Liang, Ming-Ming Cheng, Yao Zhao, and
  Shuicheng Yan.
\newblock Object region mining with adversarial erasing: A simple
  classification to semantic segmentation approach.
\newblock In {\em CVPR}, pages 1568--1576, 2017.

\bibitem{mdc}
Yunchao Wei, Huaxin Xiao, Honghui Shi, Zequn Jie, Jiashi Feng, and Thomas~S.
  Huang.
\newblock Revisiting dilated convolution: {A} simple approach for weakly- and
  semi-supervised semantic segmentation.
\newblock In {\em CVPR}, pages 7268--7277, 2018.

\bibitem{resnet38}
Zifeng Wu, Chunhua Shen, and Anton Van Den~Hengel.
\newblock Wider or deeper: Revisiting the resnet model for visual recognition.
\newblock {\em PR}, pages 119--133, 2019.

\bibitem{segformer}
Enze Xie, Wenhai Wang, Zhiding Yu, Anima Anandkumar, Jose~M Alvarez, and Ping
  Luo.
\newblock Segformer: Simple and efficient design for semantic segmentation with
  transformers.
\newblock {\em NeurIPS}, 34, 2021.

\bibitem{clims}
Jinheng Xie, Xianxu Hou, Kai Ye, and Linlin Shen.
\newblock {CLIMS:} cross language image matching for weakly supervised semantic
  segmentation.
\newblock In {\em CVPR}, 2022.

\bibitem{mctformer}
Lian Xu, Wanli Ouyang, Mohammed Bennamoun, Farid Boussa{\"{\i}}d, and Dan Xu.
\newblock Multi-class token transformer for weakly supervised semantic
  segmentation.
\newblock In {\em CVPR}, 2022.

\bibitem{bisenet}
Changqian Yu, Jingbo Wang, Chao Peng, Changxin Gao, Gang Yu, and Nong Sang.
\newblock Bisenet: Bilateral segmentation network for real-time semantic
  segmentation.
\newblock In {\em ECCV}, pages 325--341, 2018.

\bibitem{conta}
Dong Zhang, Hanwang Zhang, Jinhui Tang, Xian{-}Sheng Hua, and Qianru Sun.
\newblock Causal intervention for weakly-supervised semantic segmentation.
\newblock In {\em NeurIPS}, 2020.

\bibitem{cpn}
Fei Zhang, Chaochen Gu, Chenyue Zhang, and Yuchao Dai.
\newblock Complementary patch for weakly supervised semantic segmentation.
\newblock In {\em Proceedings of the IEEE/CVF International Conference on
  Computer Vision (ICCV)}, pages 7242--7251, 2021.

\bibitem{setr}
Sixiao Zheng, Jiachen Lu, Hengshuang Zhao, Xiatian Zhu, Zekun Luo, Yabiao Wang,
  Yanwei Fu, Jianfeng Feng, Tao Xiang, Philip~HS Torr, et~al.
\newblock Rethinking semantic segmentation from a sequence-to-sequence
  perspective with transformers.
\newblock In {\em CVPR}, pages 6881--6890, 2021.

\bibitem{cam}
Bolei Zhou, Aditya Khosla, Agata Lapedriza, Aude Oliva, and Antonio Torralba.
\newblock Learning deep features for discriminative localization.
\newblock In {\em CVPR}, pages 2921--2929, 2016.

\end{thebibliography}
}

\end{document}